\documentclass{article}
\usepackage{ijcai20}

\usepackage{times}

\usepackage{soul}
\usepackage{url}
\usepackage[hidelinks]{hyperref}
\usepackage[utf8]{inputenc}
\usepackage[small]{caption}
\usepackage{graphicx}
\usepackage{amsmath}
\usepackage{enumitem}
\usepackage{booktabs}
\usepackage{bm}
\usepackage{subcaption}
\usepackage{scalerel}
\usepackage{algorithm}
\usepackage{algorithmic}
\usepackage{listings}
\usepackage{multirow}
\usepackage[table,xcdraw]{xcolor}
\newcommand{\citet}[1]{\citeauthor{#1} \shortcite{#1}}
\newcommand{\citep}{\cite}

\usepackage{xcolor}

\begin{document}
\title{On Learning Combinatorial Patterns to Assist Large-Scale Airline Crew Pairing Optimization}

\author{
Divyam Aggarwal$^1$\and
Yash Kumar Singh$^1$\And
Dhish Kumar Saxena$^{1,}$\footnote{Contact Author}\\
\affiliations
$^1$Department of Mechanical \& Industrial Engineering, Indian Institute of Technology Roorkee, Roorkee, Uttarakhand-247667, India\\
\emails
\{daggarwal, ysingh, dhish.saxena\}@me.iitr.ac.in
}

\maketitle

\begin{abstract}
Airline Crew Pairing Optimization (CPO) aims at generating a set of \textit{legal flight sequences} (\textit{crew pairings}), to cover an airline's flight schedule, at minimum cost. It is usually performed using \textit{Column Generation} (CG), a mathematical programming technique for guided search-space exploration. CG exploits the interdependencies between the current and the preceding CG-iteration for generating new variables (pairings) during the optimization-search. However, with the unprecedented scale and complexity of the emergent flight networks, it has become imperative to \textit{learn} higher-order interdependencies among the flight-connection graphs, and utilize those to enhance the efficacy of the CPO. In first of its kind and what marks a significant departure from the state-of-the-art, this paper proposes a novel adaptation of the Variational Graph Auto-Encoder for \textit{learning} plausible combinatorial patterns among the flight-connection data obtained through the search-space exploration by an Airline Crew Pairing Optimizer, \textit{AirCROP} (developed by the authors and validated by the research consortium's industrial sponsor, \textit{GE Aviation}). The resulting flight-connection predictions are combined \textit{on-the-fly} using a novel heuristic to generate new pairings for the optimizer. The utility of the proposed approach is demonstrated on large-scale (over 4200 flights), real-world, \textit{complex} flight-networks of US-based airlines, characterized by multiple hub-and-spoke subnetworks and several crew bases.
\end{abstract}

\section{Introduction} \label{sec:intro}
Crew operating cost constitutes the second-largest component of an airline's total operating cost (next to the fuel cost). Hence, even marginal improvements in it may translate to savings of millions of dollars, annually. Given this, crew scheduling optimization assumes critical importance for any airline. \textit{Airline Crew Pairing Optimization} (CPO), being the primary step of crew scheduling, is aimed at constructing a set of legal crew pairings to cover an airline's flight schedule at minimum-cost. A \textit{legal crew pairing} is a flight sequence to be flown by an airline crew, starting and ending at the same crew base, while satisfying several complex constraints. The resulting optimization problem, referred to as CPOP, is a highly-constrained NP-Hard optimization problem \cite{bernhard2008combinatorial}. Major airlines handle large (3000+ flights) and complex flight networks on a weekly basis, resulting in billion-plus possible legal pairings. This renders their \textit{offline} enumeration intractable, and exhaustive search for their optimal full flight-coverage, impractical. A detailed review of CPOP could be found in~\cite{barnhart2003airline}.
\par CPOP is a large \textit{integer programming} (IP) problem, impractical to be solved using standard IP techniques (branch-and-bound algorithm \cite{land1960automatic}). The CPOP's literature suggests that \textit{Column Generation} (CG) is the most widely adopted optimization technique, since it allows for guided exploration of search-space based on the corresponding gain in the objective function(s). Interested readers are referred to \cite{lubbecke2010column} for an extensive review on CG. While several researchers have used CG before the IP-phase to solve CPOP's \textit{relaxed}-form, a \textit{linear programmming} (LP) problem \cite{gabriel1987column,anbil1992global,vance1997heuristic,anbil1998column}; others have used CG inside the IP-phase by adopting a \textit{branch-and-price} framework (proposed by \cite{barnhart1998branch}) \cite{desaulniers2010airline,zeren2016novel}. This research adopts the former approach, and the workflow of the optimizer is as follows. A full flight coverage set of legal pairings is used to initialize CG-phase of the optimizer. In that, several CG-iterations are performed to find a near-optimal LP-solution, which is fed in to the subsequent IP-phase. Each CG-iteration is decomposed into two problems, a \textit{restricted master problem} (RMP) and a \textit{pricing subproblem}. The RMP is modeled as a set-covering problem \cite{bernhard2008combinatorial}, and solved using the Simplex algorithm \cite{MR720547}. The dual-information embedded in the resulting solution, is then utilized to generate new pairings promising further cost-improvement (pricing subproblem). This approach suffers from two major limitations. Firstly, it may be noted that the solution to the \textit{pricing subproblem} at CG-iteration $t-1$ feeds into the RMP and \textit{pricing subproblems} at iteration $t$. Hence, effectively, CG captures the interdependencies between two subsequent iterations (local information), rendering the LP-solution vulnerable to local optimality. Secondly, there is over-reliance on the use of the dual-information. In that, any hidden/indirect flight-connection patterns which may not be obvious, yet be drivers for better pairings, have been overlooked. This paper seeks to overcome both these limitations by \textit{learning} flight-connection patterns which are recurrent from the initial till the current CG-iteration $t$, and by utilizing this knowledge to guide the the CG-\textit{search} towards otherwise unexplored (flight-connections) regions of the search space.  
\par The advancements in machine learning (ML) techniques have enabled the researchers to solve operations research (OR) problems using ML-based approaches. \citet{aytug1994areview} highlights the need to incorporate Artificial Intelligence (AI)-based methods in scheduling problems. The work on Innovization by \citet{deb2006innovization} highlights the benefits of identifying salient design rules that make a solution optimal, providing essential process knowledge to the user that may not be otherwise directly attainable. \citet{priore2014dynamic} provides a recent survey on the use of ML techniques in dynamic scheduling for manufacturing systems. \citet{wang2016applying} predicted the anticipated bus traffic by using a standard NN, however, did not use it in combination with an OR-based optimizer. A recent survey of the use of ML techniques for combinatorial optimization problems is presented in \citet{bengio2018machine}. \citet{khalil2017learning} proposed a unique combination of graph embedding and reinforcement learning techniques to learn the structure of combinatorial optimization problems which are required to be solved again and again but with different data. A similar approach has been proposed by \citet{Soumis2019FlightconnectionPF} in which regular/common flight clusters, that are part of hundreds of previous optimal solutions, are learnt. The learnt clusters are used to construct a better initial feasible solution for an airline crew pairing optimizer, drastically reducing the time to generate an acceptable sub-optimal solution. These two approaches focused only on learning regularity-based structures from multiple solutions of previous optimizer-runs which is not the aim of this research work. Convolutional Neural Networks (CNN) have proved useful in learning generalized patterns in complex structures such as image data. This concept of convolutions has been extended to graph problems, leading to the development of Graph Convolutional Networks (GCN) \citep{DBLP:journals/corr/KipfW16}. In a recent approach \cite{li2018combinatorial}, authors used the GCN to estimate the likelihood of vertices of being in the optimal solution and a guided tree search to generate a diverse set of solutions using the predicted vertices. Its utility was demonstrated for generalized and less complex combinatorial optimization problems, and that too using a GCN for semi-supervised learning which may not be useful in the context the CPOP. However, a motivation is drawn for the use of GCNs and its types for learning useful patterns in the graph-structured data.It was identified as a semi-supervised learning problem which is not useful in the context of CPOPs. As adaptive learning framework is proposed by \citet{gaur2016adaptive} to assist Multi-objective Evolutionary Algorithms. However, the same is not applicable in CPO due to its combinatorial as well as single-objective nature.
\par This research paper attempts to address the challenge of developing a \textit{learning framework} for assisting large-scale airline CPO by using a novel-adaptation of the \textit{Variational Graph Auto-encoder} (VGAE; given by \citet{Kipf2016VariationalGA}). The adapted VGAE is used to learn hidden and indirect patterns among the flight-connection graphs, constructed during CPO. A novel heuristic is proposed to combine the resulting flight-connection predictions for generating new pairings in the same CG-iteration as that of the learning. In the \textit{learning framework}:
\begin{enumerate}
\item CPOP is modeled as a graph-problem by transforming the optimization-data into a graph-structured data, enabling the adaptation of VGAE as the learning algorithm.
\item Relevant features are formulated in a way to capture the evolution of solution-quality as the optimization-search proceeds.
\item Predictions from VGAE are transformed on-the-fly into new pairings, resulting in higher cost-improvements.
\end{enumerate}
The efficacy of the proposed \textit{learning framework} is demonstrated on large-scale ($\sim$4200 flights) and complex flight networks of US-based airlines, provided by the industrial sponsor, \textit{GE Aviation}.
\par The outline of the paper is as follows: Section~\ref{sec:cpop} includes a brief discussion on the airline crew pairing problem and the developed optimizer, Section~\ref{sec:framework} describes the proposed \textit{learning framework}, Section~\ref{sec:results} presents the results of computational experiments on real-world airline flight datasets, and Section~\ref{sec:conclusion} perspectively concludes the paper.

\section{Airline Crew Pairing Optimization Problem} \label{sec:cpop}

\subsection{Terminology}
\begin{figure}[tb]
    \centering
    \includegraphics[width=0.9\columnwidth, keepaspectratio]{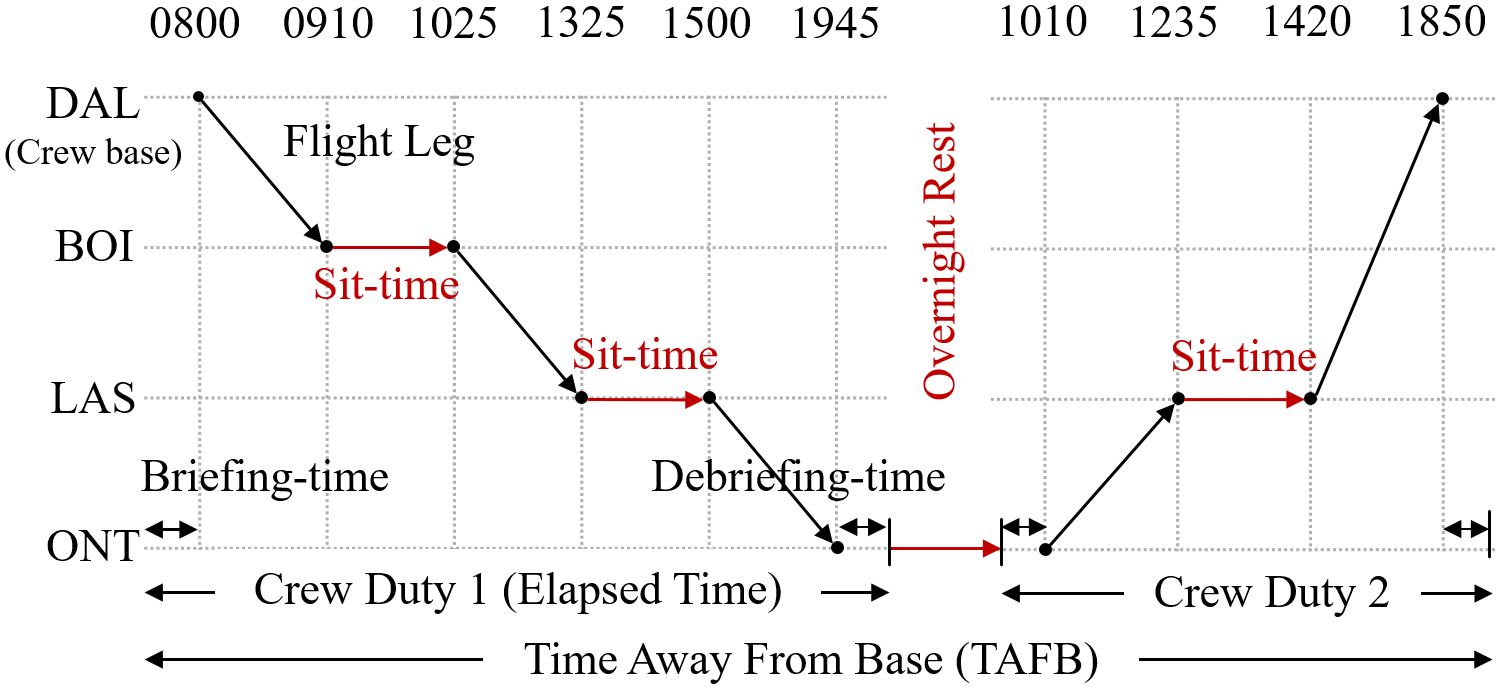}
    \caption{A legal crew pairing starting and ending at Dallas, \textit{DAL}, crew base}
    \label{fig:crewPair}
\end{figure}

A sequence of flight legs flown by a crew member, starting and ending at their home base (airport), is called a \textit{crew pairing}. Each crew is assigned a home base, called the \textit{crew base}. Each flight sub-sequence within a crew pairing that a crew flies in its working day is called a \textit{crew duty} or a \textit{duty}. Within each duty, two consecutive flights are separated by a small rest-time to allow for operations such as aircraft changes, cleaning, bagging, etc. and this is called a \textit{sit-time} or a \textit{connection-time}. A longer rest-time, provided between two consecutive duties, is called an \textit{overnight-rest}. Two shorter time-periods, \textit{briefing} and \textit{debriefing} times, are provided in the beginning and ending of a duty respectively. The total time elapsed in a crew pairing is known as the \textit{Time Away From Base} (TAFB). An example of a legal crew pairing is shown in Figure.~\ref{fig:crewPair}. In some cases, where two flights cannot be covered  in a pairing because they do not share legal flight connection, a third flight is used to connect them in which the crew travels as passenger instead of flying it. Such a flight is called a \textit{deadhead flight} for the crew going as passenger. The deadhead ing not only leads to revenue loss on the passenger seats being occupied by the deadheading crew, but also leads to paying crew wages for the deadheading hours. Hence, it is desirable for airlines to reduce deadhead flights in their crew schedules to the minimum possible, ideally zero.

\subsection{Pairing Legality Constraints \& Costing Rules}
A number of airline regulatory bodies, such as \textit{FAA}\footnote{\textit{FAA} stands for Federal Aviation Administration.}, \textit{EASA}\footnote{\textit{EASA} stands for European Aviation Safety Agency.}, etc., govern the safety of the passengers and protects the interest of the crew members. For this, several non-linear constraints have to be satisfied by a crew pairing to be classified as \textit{legal}. Moreover, several other constraints linked to an airline's in-house regulations, region-specific labor laws, etc. have to be satisfied by a legal crew pairing. Furthermore, a set of non-linear costing rules is used by airlines to calculate the cost of legal crew pairings. Interested readers are referred to \cite{aggarwal2018large} for a detailed discussion on these non-linear legality constraints and costing rules.

\subsection{Developed Airline Crew Pairing Optimizer}
The developed CG-based airline crew pairing optimizer, named as \textit{AirCROP}, is capable of generating high-quality solutions for large-scale CPOPs \citep{aggarwal2020aircrop}. It has been validated on \textit{large} and \textit{complex} flight networks (characterized by multiple hub-and-spoke subnetworks, multiple crew bases \& billion-plus legal crew pairings) in collaboration with the industrial sponsor. A number of modules make up this proprietary optimizer, namely \textit{Legal Crew Pairing Generation} \citep{aggarwal2018large}, \textit{Initial Feasible Solution Generation} \citep{aggarwal2020initializing}, and \textit{Optimization Engine}. The aim of the proposed research work is to assist the Optimization Engine module, especially in the CG-phase of the optimizer. Being a propriety optimizer, \textit{AirCROP} is used as a black-box for demonstration of the utility of this research work.

\section{Proposed Learning Framework} \label{sec:framework}
As discussed in Section~\ref{sec:intro}, the notion of incorporating domain-knowledge in CG heuristics has gained traction in solving large-scale CPOPs \cite{zeren2016novel}. 
In each CG-iteration, domain-knowledge inspired special heuristics utilize the existing flight-connection information to identify a set of critical flights. Subsequently, the identified set is used to construct the pricing subproblem from which new pairings are generated and only the pairings with negative reduced costs ($c'$) are added to the existing LP-solution, say $\mathcal{LP}$. For a pairing $p$, $c'_p$ is calculated as $c_p - \sum_{f \in p} y_{f}$, where $c_p$ is the pre-calculated cost of $p$, and $y_f$ is a dual-variable corresponding to a flight $f$. For a $t^{th}$ CG-iteration, these dual-variables are calculated by solving a dual-model of the CPOP. This dual-model is constructed using $\mathcal{LP}^{(t)}$ and the given flight set, $\mathcal{F}$. The dual-variables are collectively represented by a dual-vector, $\mathbf{y}^{(t)}$, which is a column-matrix, given as $\bm{[}\!\!\bm{[}y_{f}^{(t)}\ \forall f\in \mathcal{F}\ |\ y_{f}^{(t)}\geq 0\bm{]}\!\!\bm{]}^{\mathsf{T}}$. In this manner, the interdependencies between the active ($t^{th}$) and its immediately preceding iteration ($(t-1)^{th}$) are captured.
\par In CPOP, being a combinatorial optimization problem, a pairing (optimization-variable) is neither explicitly good nor explicitly bad until evaluated in a feasible set of pairings. At a $t^{th}$ CG-iteration, it is highly likely that a pairing rejected from any of the previous LP-solutions ($\mathcal{LP}^{(t-1)}, \mathcal{LP}^{(t-2)},...,or \mathcal{LP}^{(1)}$), due to its lower contribution in them, might become a critical-part of the $\mathcal{LP}^{(t)}$. Similarly, there might exist other implicit relationships which are not interpretable using the existing domain-knowledge. This builds the rationale for the development of the proposed \textit{learning framework} that attempts to learn implicit combinatorial pattern(s), hidden in the optimization-data. In the proposed framework, the learning is performed at the end of multiple CG-iterations, each called a \textit{learning-iteration}. The learnt pattern(s) are simultaneously used (in the same iteration) to assist the convergence of the succeeding iterations. The solution's quality is extremely poor in initial CG-iterations and improves as the optimization-search converges. Hence, to avoid pre-mature learning, the learning-iterations are kept at larger-gaps initially. The frequency of learning is increased as the optimization-search converges. What happens in each of these learning-iterations is discussed in detail in the following subsections. 

\subsection{Input Dataset}
\begin{figure}[htb]
    \centering
    \includegraphics[width=0.75\columnwidth, keepaspectratio]{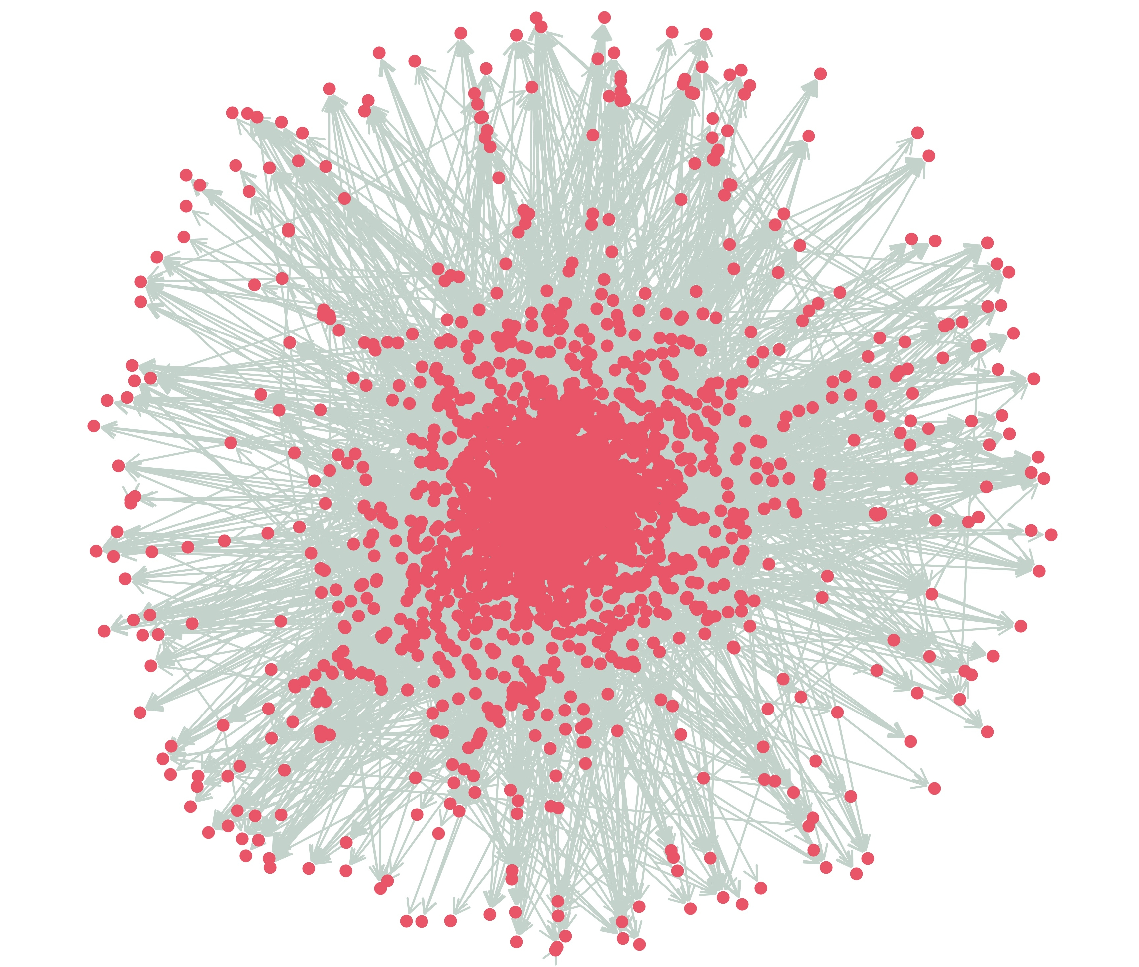}
    \caption{Flight-connection graph of a small $\mathcal{LP}$ for a CPOP, covering 3202 flight nodes using 21,564 edges in 7991 pairings}
    \label{fig:connectedFltGraph}
\end{figure}

At a $t^{th}$ CG-iteration, the optimization-data from all $t$ CG-iterations is available. This consists of:
\begin{itemize}[leftmargin=*]
\item LP-solutions ($\mathcal{LP}^{(t)},\mathcal{LP}^{(t-1)},...\mathcal{LP}^{(1)}$),
\item cost of LP-solutions ($C^{(t)},C^{(t-1)},...C^{(1)}$), and
\item dual-vectors ($\mathbf{y}^{(t)}, \mathbf{y}^{(t-1)},...\mathbf{y}^{(1)}$)
\end{itemize}
An $\mathcal{LP}^{(t)}$ consists of a set of pairings $\mathcal{P}^{(t)}$ and its primal-vector $\mathbf{x}^{(t)}$ which is  a column-matrix, given as $\bm{[}\!\!\bm{[}x_{p}^{(t)}\ \forall p \in \mathcal{P}^{(t)}\ |\ 0 \leq x_{p}^{(t)} \leq 1\bm{]}\!\!\bm{]}^{\mathsf{T}}$. A $\mathbf{x}^{(t)}$ is simply a weight-vector of $\mathcal{P}^{(t)}$, representing relative-contribution of its pairings towards covering the flights in $\mathcal{F}$. During experiments, it is observed that the optimization-search of a large and complex CPOP (4212 flights and 15 crew bases) involves generation and evaluation of billion-plus pairings. This makes it impractical to explicitly use these billion-plus pairings as input data points for the learning algorithm. Alternatively, $\mathcal{LP}^{(t)}$ is modeled as a weighted directed multigraph $\mathcal{G}^{(t)}$, 
given by an ordered pair $(\mathcal{F},\mathcal{E}^{(t)})$, where flights in $\mathcal{F}$ become vertices, and $\mathcal{E}^{(t)}$ is the set of directed edges (legal flight-connection), given as $\{(f_i,f_j)\ |\ (f_i,f_j) \in p_k\ \forall\ p_k \in \mathcal{P}^{(t)}\ \land\ i<j\}$. Please note that $\mathcal{G}$ is referred to as \textit{flight-connection graph} interchangeably throughout this paper. In the test-instances used in this research work, the flight indexes are sorted with-respect-to their timestamps\footnote{A \textit{timestamp} gives the information about the date and time of the event.}. And a flight cannot have a legal flight connection with itself. Hence, the ordered pairs in $\{(f_i,f_j)\ |\ (f_i,f_j)\in \mathcal{F}^2 \land i>=j\}$ are invalid and excluded from the definition of $\mathcal{G}^{(t)}$. Figure~\ref{fig:connectedFltGraph} presents a flight-connection graph of a very small pairing set for a CPOP with 3202 flights. 
The largest test-case provided by the industrial sponsor contains $4212$ flights and its number of all legal flight connections will be $\leq 4212 \times 4212$. Hence, the adjacency matrix $\mathbf{A}^{(t)}$ (size $|\mathcal{F}|\times |\mathcal{F}|$) of a flight-connection graph $\mathcal{G}^{(t)}$ is a viable representation of an input data point than a huge set of pairings (matrix of size $|\mathcal{P}|\times |\mathcal{F}|$ where $|\mathcal{P}|$ can vary upto a million). Therefore, a \textit{global adjacency matrix} $\mathbf{\tilde{A}}^{(t)}$ is constructed to be used as input dataset for the learning algorithm. In addition to this, an \textit{input feature matrix} $\mathbf{F}^{(t)}$ is constructed by formulating critical features from the available optimization-data. These two matrices are discussed in detail in the following text.
\subsubsection{Global Adjacency Matrix}
For a flight-connection graph $\mathcal{G}^{(t)}$ with vertex set $\mathcal{F}$, the adjacency matrix $\mathbf{A}^{(t)}$ is a square matrix of size $|\mathcal{F}|\times |\mathcal{F}|$ whose elements represent the legal flight-connections, i.e., edges in $\mathcal{E}^{(t)}$. 
An element $a_{ij}^{(t)}$ of $\mathbf{A}^{(t)}$, corresponding to its $i^{th}$ row and $j^{th}$ column, is a binary variable, and is either $=1$ if the ordered pair $(f_i,f_j)$ exists in $\mathcal{G}^{(t)}$, or is $=0$ otherwise. In accordance with the definition of $\mathcal{G}^{(t)}$, matrix $\mathbf{A}^{(t)}$ is a strictly upper-triangular matrix. 
\par For learning at a $t^{th}$ CG-iteration, a \textit{global adjacency matrix} $\mathbf{\tilde{A}}^{(t)}$ is constructed by superimposing individual adjacency matrices $\mathbf{A}^{(k)}\ \forall k\ \in [1,t]$. Let $\tilde{a}^{(t)}_{ij}$ be the element of matrix $\mathbf{\tilde{A}}^{(t)}$, corresponding to its $i^{th}$ row and $j^{th}$ column. Therefore,
\begin{equation}
    \tilde{a}^{(t)}_{ij}  =
    \begin{cases}
      1, & \text{if}\ a_{ij} = 1\ \in \mathbf{A}^{(1)}\cup\mathbf{A}^{(2)}\cup...\cup\mathbf{A}^{(t)}\\
      0, & \text{otherwise}
    \end{cases}
\end{equation}


\subsubsection{Feature Matrix}
For higher prediction accuracy of any learning algorithm, it is desirable to train it not only on a large number of features but also on the most critical ones. From the available optimization-data, several critical features are formulated using the domain-knowledge. As mentioned before, the solution quality (LP-cost, $C^{(t)}$) improves iteratively during the optimization-search and it is imperative to ensure that the characteristics of a $t^{th}$ CG-iteration are weighed more than those of the $(t-1)^{th}$ iteration. This is done by multiplying them with a \textit{cost-ratio}, $CR^{(t)}$, given as $CR^{(t)} = C^{(t-1)}/C^{(t)}$. This enables the learning algorithm to capture the solution's evolution by giving higher weights to the better and more recent solution. The features formulated using the primal- and dual-information, available in the optimization-data at a $t^{th}$ iteration, are as follows:
\begin{enumerate}[label=\textbf{F\arabic*-}]
\item \textbf{Enhanced Primal Matrix, $\mathbf{X}^{(t)}$:} As explained above, in a $t^{th}$ CG-iteration, the fractional contribution of a pairing $p \in \mathcal{P}^{(t)}$ for covering a flight-connection $(f_i,f_j) \in \mathcal{G}^{(t)}$ is given by its primal-variable $x_p^{(t)}$. Hence, the weight $w_{ij}^{(t)}$ of an edge $(f_i,f_j)$ is given as $w_{ij}^{(t)}=\sum_{p \in \mathcal{P}^{(t)}} (x_p^{(t)}\ |\ (f_i,f_j) \in p)$. Using these weights and matrix $\mathbf{A}^{(t)}$ , a weighted-adjacency matrix $\mathbf{\hat{A}}^{(t)}$ is defined whose elements $\hat{a}^{(t)}_{ij}$ are given as $\hat{a}^{(t)}_{ij} = a^{(t)}_{ij}*w_{ij}^{(t)}$. The $\mathbf{\hat{A}}^{(t)}$ for all $t$ iterations are enhanced using the corresponding $CR^{(t)}$ and are added to construct an \textit{enhanced primal matrix} $\mathbf{X}^{(t)}$ of size $|\mathcal{F}|\times |\mathcal{F}|$. Therefore, $\mathbf{X}^{(t)} = CR^{(1)}.\mathbf{\hat{A}}^{(1)}+CR^{(2)}.\mathbf{\hat{A}}^{(2)}+...+CR^{(t)}.\mathbf{\hat{A}}^{(t)}$. It is to be noted that $CR^{(1)}$ is assumed to be $=1$ on the basis that $C^{(0)}$ does not exists.

\item \textbf{Enhanced Dual Matrix, $\mathbf{Y}^{(t)}$:} In a $t^{th}$ CG-iteration, a dual-vector $\mathbf{y}^{(t)}$ consists of dual-variables which are shadow-prices of the corresponding flights and reflect their contribution in the $C^{(t)}$. The distribution of these fractional contributions ($y_f^{(t)}/C^{(t)}$) over all $t$ iterations for all flights does not follow the same trend and are not explicitly interpretable. Hence, these flights' fractional contributions for each iteration are concatenated to construct an \textit{enhanced dual matrix} $\mathbf{Y}^{(t)}$ of size $|\mathcal{F}|\times t$, given as\\$\mathbf{Y}^{(t)} = \bm{[}\!\!\bm{[}\frac{CR^{(1)}}{C^{(1)}}.\mathbf{y}^{(1)}\ \ \ \ \ \ \frac{CR^{(2)}}{C^{(2)}}.\mathbf{y}^{(2)}\ \ \ \ \ \ ...\ \ \ \ \ \ \frac{CR^{(t)}}{C^{(t)}}.\mathbf{y}^{(t)}\bm{]}\!\!\bm{]}$.
\end{enumerate}
The features formulated using the graphical properties of the flight nodes are as follows:
\begin{enumerate}[label=\textbf{F\arabic*-},start=3]
\item \textbf{Enhanced In-degree Matrix, $\mathbf{\tilde{I}}^{(t)}$:} In a flight-connection graph, the sum of weights of incoming edges for a flight $f$ is referred to as its \textit{in-degree}, denoted by $\text{deg}^{-}(f)$. For a $\mathcal{G}^{(t)}$, an in-degree matrix $\mathbf{\hat{I}}^{(t)}$ is constructed whose elements are $=sum(\hat{a}_{*,j})$, where $\hat{a}_{*,j}$ is the $j^{th}$ column of matrix $\mathbf{\hat{A}}^{(t)}$. Using these $\mathbf{\hat{I}}^{(t)}$ and the corresponding $CR^{(t)}$, an \textit{enhanced in-degree matrix} $\mathbf{\tilde{I}^{(t)}}$ of size $|\mathcal{F}|\times 1$ is constructed which is given as $\mathbf{\tilde{I}^{(t)}}= CR^{(1)}.\mathbf{\hat{I}}^{(1)}+CR^{(2)}.\mathbf{\hat{I}}^{(2)}+...+CR^{(t)}.\mathbf{\hat{I}}^{(t)}$.
\item \textbf{Enhanced Out-degree Matrix, $\mathbf{\tilde{O}}^{(t)}$:} In a flight-connection graph, the sum of weights of outgoing edges from a flight $f$ is referred to as its \textit{out-degree}, denoted by $\text{deg}^{+}(f)$. For a $\mathcal{G}^{(t)}$, an out-degree matrix $\mathbf{\hat{O}}^{(t)}$ is constructed whose elements are $=sum(\hat{a}_{i,*})$, where $\hat{a}_{i,*}$ is the $i^{th}$ row of matrix $\mathbf{\hat{A}}^{(t)}$. Using these $\mathbf{\hat{O}}^{(t)}$ and the corresponding $CR^{(t)}$, an \textit{enhanced out-degree matrix} $\mathbf{\tilde{O}}^{(t)}$ of size $|\mathcal{F}|\times 1$ is constructed which is given as $\mathbf{\tilde{O}}^{(t)}= CR^{(1)}.\mathbf{\hat{O}}^{(1)}+CR^{(2)}.\mathbf{\hat{O}}^{(2)}+...+CR^{(t)}.\mathbf{\hat{O}}^{(t)}$.
\end{enumerate}

These two features help gauge the importance of flights by measuring how densely each flight is connected to other flights. The flight(s) with higher in-degree and out-degree might be intuitively considered to be important than the other flights.
\par Each of the above-defined features is normalized within itself using a min-max normalization technique \cite{hastie2009elements} so that the values are scaled to $[0,1]$. This ensures that none of the features get unduly high weightage and all features are on a similar scale. After feature re-scaling, these feature matrices are concatenated to construct the \textit{input feature matrix} $\mathbf{F}^{(t)}$, given as $\mathbf{F}^{(t)} = \bm{[}\!\!\bm{[} \mathbf{X}^{(t)}_{\scaleto{|\mathcal{F}|\times |\mathcal{F}|}{5pt}} \ \ \ \mathbf{Y}^{(t)}_{\scaleto{|\mathcal{F}|\times t}{5pt}} \ \ \ \mathbf{\tilde{I}}^{(t)}_{\scaleto{|\mathcal{F}|\times 1}{5pt}} \ \ \ \mathbf{\tilde{O}}^{(t)}_{\scaleto{|\mathcal{F}|\times 1}{5pt}} \bm{]}\!\!\bm{]}$, and its size is $|\mathcal{F}|\times N$ where $N = |\mathcal{F}|+t+2$.

\subsection{Learning Algorithm}
\par
\begin{figure}[b]
    \centering
    \includegraphics[width=0.4\textwidth, keepaspectratio]{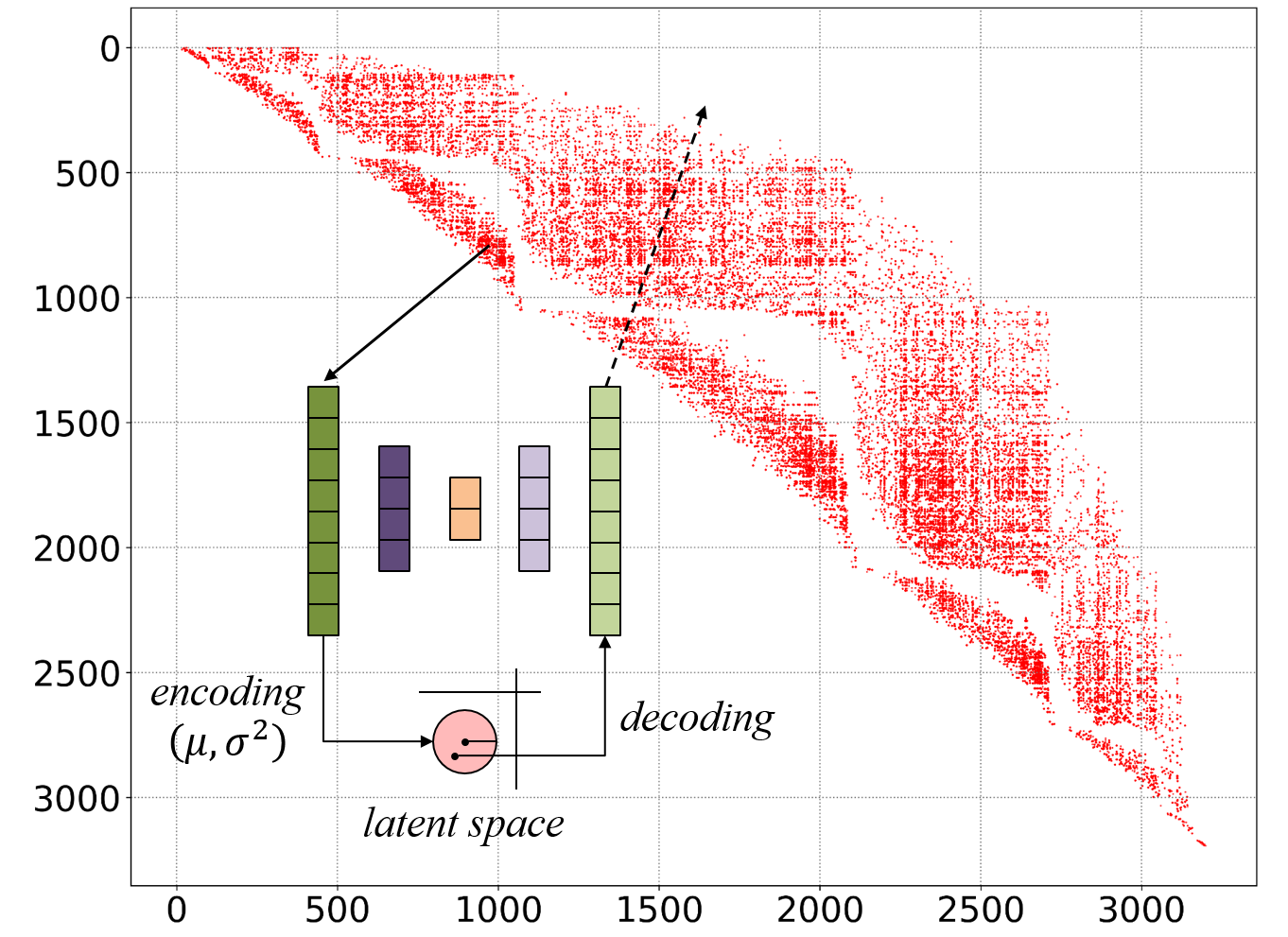}
    \caption{Working of VGAE}
    \label{fig:working}
\end{figure}

 In the proposed framework, a \textit{Variational Graph Autoencoder} (VGAE) \cite{Kipf2016VariationalGA} is adapted as the learning algorithm. VGAE is a type of variational auto-encoders, developed for unsupervised learning on graph-structured data by using the interpretable latent representations of the graphs. \cite{Kipf2016VariationalGA} demonstrated the supremacy of VGAE over traditional methods of link prediction and unsupervised learning on graph-structured data. A working representation of VGAE in the context of airline CPOP is demonstrated in Figure~\ref{fig:working}.
The figure represents the input dataset of the learning algorithm, i.e., matrix $\mathbf{\tilde{A}}^{(t)}$ at a $t^{th}$ CG-iteration of an optimizer-run for a 3202 flight dataset. Let $\mathcal{\tilde{E}}^{(t)+}$ \& $\mathcal{\tilde{E}}^{(t)-}$ be the disjoint sets of flight-pairs corresponding to all $1$'s \& all $0$'s in $\mathbf{\tilde{A}}^{(t)}$ respectively. Hence, $\mathcal{\tilde{E}}^{(t)+} = \{(f_i,f_j)|(f_i,f_j) \in \mathbf{\tilde{A}}^{(t)}\land \tilde{a}^{(t)}_{ij}=1\}$, and $\mathcal{\tilde{E}}^{(t)-} = \{(f_i,f_j)|(f_i,f_j) \in \mathbf{\tilde{A}}^{(t)}\land \tilde{a}^{(t)}_{ij}=0\}$. Let $\mathcal{M}^{(t)}_l$ be the learning model. The output of $\mathcal{M}^{(t)}_l$ includes the prediction values, $P(\mathcal{\tilde{E}}^{(t)+})\ \& \ P(\mathcal{\tilde{E}}^{(t)-})$, and its ROC score, $roc(\mathcal{M}^{(t)}_l)$ in decimal. The aim of the proposed learning algorithm is not only to find the most convoluted flight-pairs but also to find a set of flight-pairs different from $\mathcal{\tilde{E}}^{(t)+}$. This helps in expanding the search-space of the optimizer by generating pairings different from the ones encountered during the optimization-search. Hence, the only relevant information from the output of $\mathcal{M}^{(t)}_l$ includes $roc(\mathcal{M}^{(t)}_l)$ and $P(\mathcal{\tilde{E}}^{(t)-})$.

\subsection{Integration with \textit{AirCROP}}
The proposed \textit{learning framework} is aimed to assist in the CG-iterations of \textit{AirCROP} by providing a set of most critical flight-pairs, after learning from previous solutions. These learnt flight-pairs are combined using a special heuristic to generate new legal pairings. The pseudo code of this combination heuristic is summarized in Algorithm~\ref{alg:heuristic}.
\begin{algorithm}[tb]
\caption{Combination heuristic}
\label{alg:heuristic}
\textbf{Input}: $\mathcal{\tilde{E}}^{(t)-}, P(\mathcal{\tilde{E}}^{(t)-}), Acc(\mathcal{M}^{(t)}_l), \mathcal{F}, \mathbf{y}^{(t)}, \verb|Pairing_Gen()|$\\
\textbf{Parameter}: $Param_1$\\
\textbf{Output}: $\mathcal{\tilde{P}}^{(t)}$
\begin{algorithmic}[1] 
\STATE $\mathcal{\tilde{F}}^{(t)}\leftarrow\ \phi$
\STATE $\mathcal{\tilde{E}}^{(t)-}_{s}\leftarrow\ \verb|sort_descend(|\mathcal{\tilde{E}}^{(t)-}\verb|)|$ using $P(\mathcal{\tilde{E}}^{(t)-})$
\FOR{$(f_i,f_j) \in \mathcal{\tilde{E}}^{(t)-}_{sort}$}
\IF {$|\mathcal{\tilde{F}}^{(t)}| \leq \lfloor Param_1* Acc(\mathcal{M}^{(t)}_l)\rfloor$}
\STATE $\mathcal{\tilde{F}}^{(t)}\leftarrow\ $Add flights $f_i$ \& $f_j$
\ELSE
\STATE break
\ENDIF
\ENDFOR
\STATE $\gamma\leftarrow\ Param_1 - |\mathcal{\tilde{F}}^{(t)}|$
\STATE $\mathcal{\tilde{F}}^{(t)}\leftarrow\ $Randomly select $\gamma$ flights from $\mathcal{F}$ 
\STATE $\mathcal{\tilde{P}}^{(t)}\leftarrow\ \verb|Pairing_Gen(|\mathcal{\tilde{F}}^{(t)}, \mathbf{y}^{(t)}\verb|)|$
\STATE \textbf{return} $\mathcal{\tilde{P}}^{(t)}$
\end{algorithmic}
\end{algorithm}
Let $\mathcal{\tilde{F}}^{(t)}$ \& $\mathcal{\tilde{P}}^{(t)}$ be the set of flights to be extracted from the learnt flight-pairs and the set of pairings to be generated from these flights. Let $Param_1$ be a user-defined parameter which decides the number of flights to be added to $\mathcal{\tilde{F}}^{(t)}$. It is restricted to be $< |\mathcal{F}|/2$ to maintain tractability in the subsequent pairing generation phase. In line 2, $\mathcal{\tilde{E}}^{(t)-}$ is sort in descending-order w.r.t. $P(\mathcal{\tilde{E}}^{(t)-})$. In lines 3-9, individual flights from the top flight-pairs of $\mathcal{\tilde{E}}^{(t)-}_{s}$ are added to $\mathcal{\tilde{F}}^{(t)}$ until its length exceeds a user-defined limit $\lfloor Param_1* roc(\mathcal{M}^{(t)}_l)\rfloor$. The $roc(\mathcal{M}^{(t)}_l)$ is used here to account for unreliability factor ($1- roc(\mathcal{M}^{(t)}_l)$) in the results of $\mathcal{M}^{(t)}_l$. The rest of the flights ($Param_1 - \lfloor Param_1* roc(\mathcal{M}^{(t)}_l)\rfloor$) are selected randomly from $\mathcal{F}$. Finally, in line 12, the pairing generation module is called using $\mathcal{\tilde{F}}^{(t)}$ \& $\mathbf{y}^{(t)}$ as inputs, to generate new legal pairings with negative reduced costs. As discussed in Section~\ref{sec:framework}, the \textit{learning framework} performs learning several times during the CG-iterations of \textit{AirCROP}. The decision on \textit{how much gap to maintain between two successive learning-iterations} and \textit{whether the gap should be constant or adaptive}, is left in the hands of the airline users to cater to their varying requirements.

\begin{table}[t]
\centering
\resizebox{0.95\columnwidth}{!}{%
\begin{tabular}{l|rr|rr|rr}  
\toprule
\multirow{2.5}{*}{$n$} & \multicolumn{2}{c|}{$\mathbf{\alpha}=0.001$} & \multicolumn{2}{c|}{$\mathbf{\alpha}=0.01$} & \multicolumn{2}{c}{$\mathbf{\alpha}=0.1$}\\
\cmidrule{2-3} \cmidrule{4-5} \cmidrule{6-7}
& $roc$ & $t\ (s)$ & $roc$ & $t\ (s)$ & $roc$ & $t\ (s)$ \\
\midrule
100 & 0.75& 219.97&\textbf{0.90}& \textbf{214.68}& 0.72& 213.07\\
\midrule
500 & 0.85& 1050.75& 0.94& 1053.98& 0.71 & 1054.05\\
\midrule
1000 & 0.91& 2124.94& 0.94& 2108.37& 0.50& 2093.39\\
\bottomrule
\end{tabular}}
\caption{First step of hyperparameter tuning}
\label{tab:tuning1}
\end{table}
\section{Computational Experiments} \label{sec:results}
All computational experiments have been conducted on a HP-Z640 workstation, powered by Intel\textsuperscript{\textregistered} Xeon\textsuperscript{\textregistered} Processor E5-2630v3, with 64GB of RAM and 32 cores at 2.4 GHz clock speed. The proposed methodology is implemented using \textit{python} v3.6 programming language. The VGAE has been adapted from its original code, given by \cite{Kipf2016VariationalGA}. The internal settings of \textit{AirCROP} are kept constant for all runs so that the results are not skewed and are comparable. The run-time of \textit{AirCROP} (without the \textit{learning framework}) is around 6 \& 14 hours for CPOPs with around 3200 \& 4200 flights respectively. To keep the run-time of \textit{AirCROP} with \textit{learning framework} as low as possible, the learning model is allowed to terminate as soon as $roc(\mathcal{M}_l^{(t)}) \geq 0.9$.
\par For this research work, the industrial sponsor has provided three real-world large-scale airline test-cases,  (1) \textit{TC1} with 3202 flights, (2) \textit{TC2} with 3228 flights, and (3) \textit{TC3} with 4212 flights, from the networks of their client airlines. These test-cases, extracted from networks of US-based airlines, contain multiple hub-and-spoke subnetworks and 15 crew bases. As a result, the number of possible legal pairings is extremely large (in the order of millions/billions). A set of pairing's legality constraints and costing rules are also provided in these test-cases. An interesting observation is that it is these legality constraints which are responsible for the shape of the flight-connections in Figure~\ref{fig:working}.

\begin{table}[b]
\centering
\resizebox{0.77\columnwidth}{!}{%
\begin{tabular}{l|rrrr}  
\toprule
 \multirow{2.5}{*}{Mesure(s)} & \multicolumn{4}{c}{$\alpha$}\\
\cmidrule{2-5}
 & $0.02$ & $0.03$ & $0.04$ & $0.05$\\
\midrule
$roc$ & 0.901& \textbf{0.919}& 0.877& 0.861\\
$t\ (s)$ & 215.79& \textbf{213.32}& 215.32& 214.64\\
\bottomrule
\end{tabular}}
\caption{Second step of hyperparameter tuning ($n=100$)}
\label{tab:tuning2}
\end{table}
\begin{table*}[tb]
\resizebox{\textwidth}{!}{%
\begin{tabular}{ll|rrrrr|rrrrr|rrrrr}
\toprule
\multicolumn{2}{c|}{\multirow{3}{*}{AirCROP's}} & \multicolumn{5}{c|}{TC1}                                                                                                                                   & \multicolumn{5}{c|}{TC2}                                                                                                                                   & \multicolumn{5}{c}{TC3}                                                                                                                                   \\
\cmidrule{3-17}
\multicolumn{2}{c|}{\multirow{4}{*}{Modules}}  & \multicolumn{2}{c}{w/o Learning} & \multicolumn{2}{c}{w/ Learning} & \multirow{1.5}{*}{Change}& \multicolumn{2}{c}{w/o Learning} & \multicolumn{2}{c}{w/ Learning} & \multirow{1.5}{*}{Change}& \multicolumn{2}{c}{w/o Learning} & \multicolumn{2}{c}{w/ Learning} & \multirow{1.5}{*}{Change}
\\
\cmidrule{3-4}\cmidrule{5-6}\cmidrule{8-9}\cmidrule{10-11}\cmidrule{13-14}\cmidrule{15-16}
                                                            &    & Cost         & \multirow{2}{*}{z}           & Cost & \multirow{2}{*}{z}          & \multirow{-1.25}{*}{in cost} & Cost & \multirow{2}{*}{z}           & Cost         & \multirow{2}{*}{z}          & \multirow{-1.25}{*}{in cost}                                                                                     & Cost         & \multirow{2}{*}{z}           & Cost         & \multirow{2}{*}{z}          & \multirow{-1.25}{*}{in cost}                                                                                    
\\
                                                            &    &(USD)         &           &(USD)         &          &(USD)                                                                                      &(USD)         &           &(USD)         &          &(USD)                                                                                     &(USD)         &           &(USD)         &         &(USD)
\\\midrule
\multicolumn{2}{l|}{Initial Solution}                             & 4,696,619            & 01          & 4,696,619            & 01         & 0                                                                                    & 5,002,751            & 01          & 5,002,751            & 01         & 0                                                                                    & 5,415,114            & 01          & 5,415,114            & 01         & 0                                                                                    \\\midrule
Main-Opt & CG & 3,475,707            & 65          & 3,468,115            & 79         & \cellcolor[HTML]{CEEECD}-7,592                                                         & 3,499,950            & 63          & 3,499,832            & 75         & \cellcolor[HTML]{CEEECD}-118                                                          & 4,599,803            & 134         & 4,592,395            & 131        & \cellcolor[HTML]{CEEECD}-7,408                                                         \\
Loop   & IP & 3,685,633            & 01          & 3,710,720            & 01         & \cellcolor[HTML]{FFCCC9}+25,087                                                       & 3,748,000            & 01          & 3,692,650            & 01         & \cellcolor[HTML]{CEEECD}-55,350                                                        & 4,896,974            & 01          & 4,935,672            & 01         & \cellcolor[HTML]{FFCCC9}+38,698                                                       \\\midrule
Re-Opt & CG & 3,483,479            & 28          & 3,472,723            & 31         & \cellcolor[HTML]{CEEECD}-10,756                                                        & 3,505,069            & 38          & 3,504,403            & 34         & \cellcolor[HTML]{CEEECD}-666                                                          & 4,613,596            & 50          & 4,6105,94            & 67         & \cellcolor[HTML]{CEEECD}-3,002                                                         \\
Loop 1                                                            & IP & 3,581,795            & 01          & 3,565,269            & 01         & \cellcolor[HTML]{CEEECD}-16,526                                                        & 3,559,727            & 01          & 3,545,146            & 01         & \cellcolor[HTML]{CEEECD}-14,581                                                        & 4,710,090            & 01          & 4,739,417            & 01         & \cellcolor[HTML]{FFCCC9}+29,327                                                       \\\midrule
Re-Opt   & CG & 3,487,314            & 22          & 3,476,249            & 24         & \cellcolor[HTML]{CEEECD}-11,065                                                        & 3,513,929            & 17          & 3,509,138            & 25         & \cellcolor[HTML]{CEEECD}-4,791                                                         & 4,632,585            & 36          & 4,634,389            & 47         & \cellcolor[HTML]{FFCCC9}+1,804                                                        \\
Loop 2                                                            & IP & 3,541,958            & 01          & 3,509,156            & 01         & \cellcolor[HTML]{CEEECD}-32,802                                                        & 3,528,128            & 01          & 3,526,182            & 01         & \cellcolor[HTML]{CEEECD}-1,946                                                         & 4,652,460            & 01          & 4,642,060            & 01         & \cellcolor[HTML]{CEEECD}+10,400                                                        \\\midrule
Re-Opt    & CG & 3,487,476            & 22          & 3,480,835            & 21         & \cellcolor[HTML]{CEEECD}-6,641                                                         & 3,513,784            & 17          & 3,509,916            & 19         & \cellcolor[HTML]{CEEECD}-3,868                                                         & 4,631,449            & 35          & 4,632,915            & 26         & \cellcolor[HTML]{FFCCC9}+1,466                                                        \\
Loop 3                                                            & IP & 3,511,944            & 01          & 3,501,972            & 01         & \cellcolor[HTML]{CEEECD}-9,972                                                         & 3,513,784            & 01          & 3,512,760            & 01         & \cellcolor[HTML]{CEEECD}-1,024                                                         & 4,648,552            & 01          & 4,637,990            & 01         & \cellcolor[HTML]{CEEECD}-10,562                                                        \\\midrule
Re-Opt    & CG & 3,487,662            & 20          & 3,482,579            & 14         & \cellcolor[HTML]{CEEECD}-5,083                                                         & 3,513,437            & 12          & 3,509,986            & 18         & \cellcolor[HTML]{CEEECD}-3,451                                                         & 4,631,700            & 26          & 4,632,343            & 27         & \cellcolor[HTML]{FFCCC9}+643                                                         \\
Loop 4                                                            & IP & 3,502,514            & 01          & 3,491,307 & 01         & \cellcolor[HTML]{CEEECD}-11,207 & 3,513,437            & 01          & 3,513,232            & 01         & \cellcolor[HTML]{CEEECD}-205 & 4,639,736            & 01          & 4,632,343            & 01         & \cellcolor[HTML]{CEEECD}-7,393                                                      \\\midrule
Re-Opt    & CG &                    &             &                    &            &                                                                                      &                    &             &                    &            &                                                                                      & 4,634,794            & 17          &                    &            &                                                                                      \\
Loop 5                                                            & IP &                    &             &                    &            &                                                                                      &                    &             &                    &            &                                                                                      & 4,634,794            & 01          &                    &            &                                                                                      \\
\midrule
\multicolumn{2}{l|}{Final Results}                                & 3,502,514            & 163         & \textbf{3,491,307}            & 175        & \textbf{11,207}                                                                                & 3,513,437            & 153         & \textbf{3,513,232}            & 177        & \textbf{205}                                                                                  & 4,634,794            & 287         & \textbf{4,632,343}            & 304        & \textbf{2,451}                                                                                 \\\midrule
\multicolumn{2}{l|}{Time (hrs)}                                   & \multicolumn{2}{r}{6.53}         & \multicolumn{2}{r}{13.14}       &                                                                                      & \multicolumn{2}{r}{5.82}         & \multicolumn{2}{r}{13.43}       &                                                                                      & \multicolumn{2}{r}{13.58}        & \multicolumn{2}{r}{29.69}       & \\
\bottomrule
\end{tabular}}
\caption{Results of \textit{AirCROP}'s runs with \& without the \textit{learning framework}}
\label{tab:runs}
\end{table*}

\begin{figure*}[tb]
\centering
\begin{subfigure}[t]{0.33\textwidth}
    \centering
    \includegraphics[width=\textwidth, keepaspectratio]{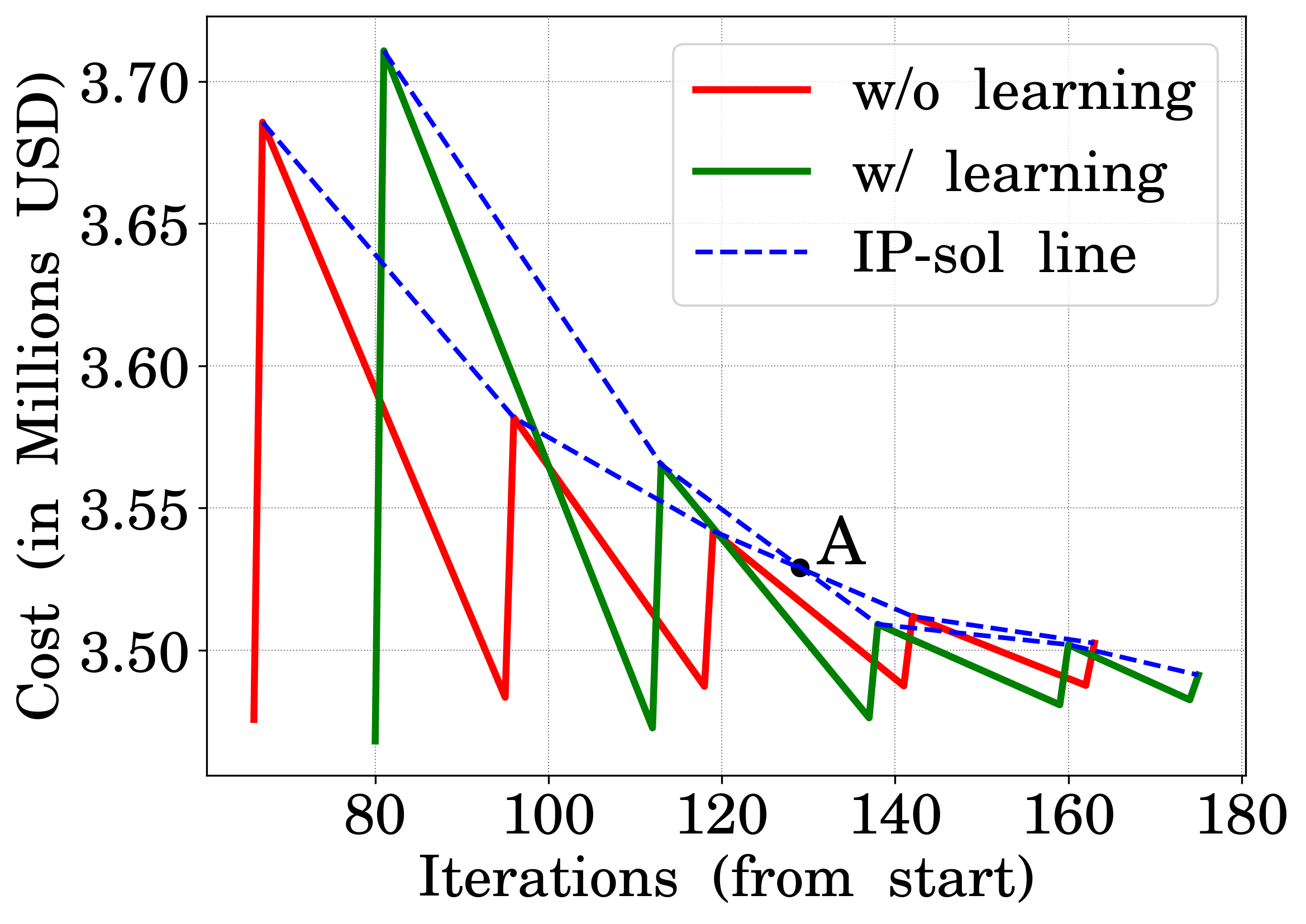}
    \caption{TC1}
    \label{fig:tc1}
\end{subfigure}
\hfill
\begin{subfigure}[t]{0.33\textwidth}
    \centering
    \includegraphics[width=\textwidth, keepaspectratio]{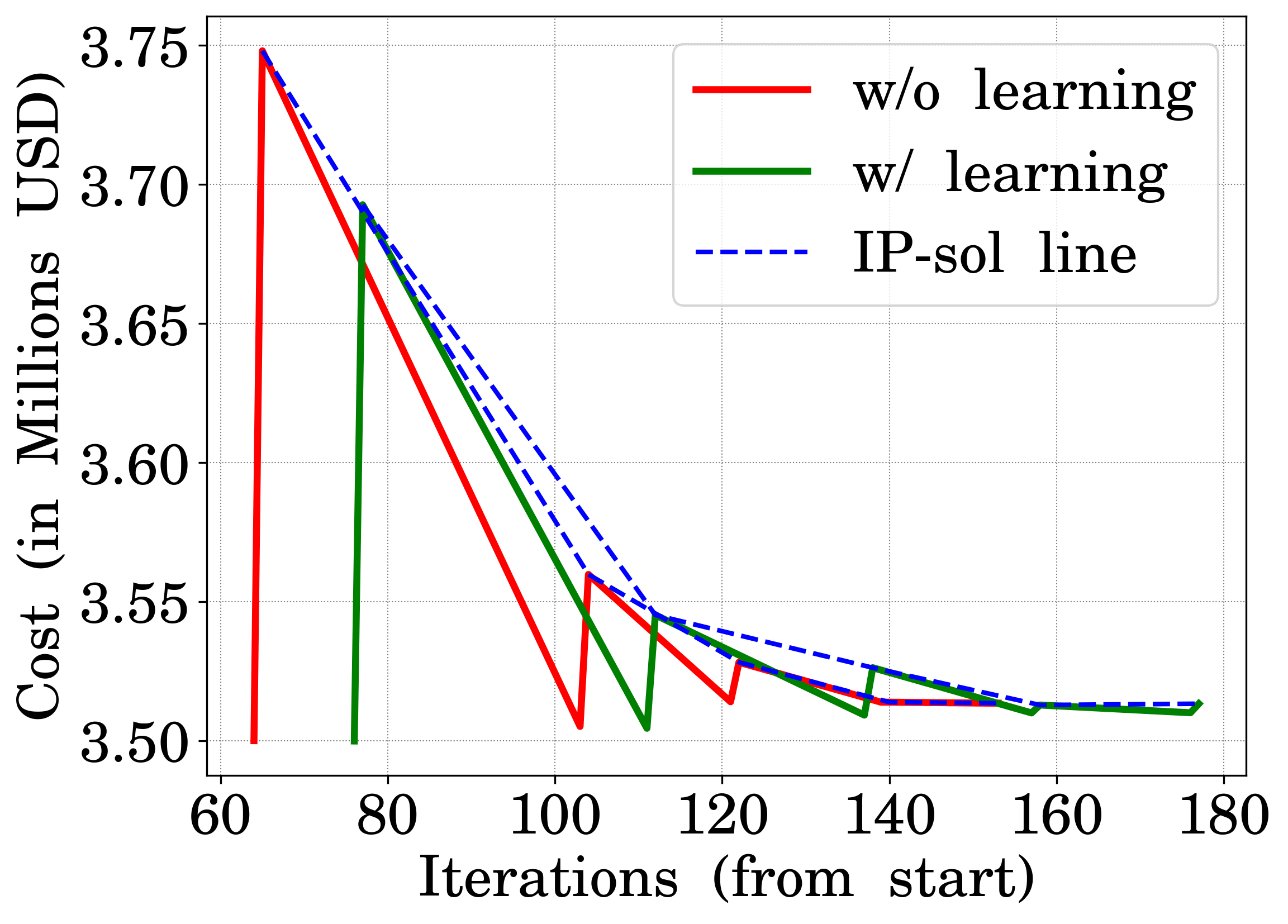}
    \caption{TC2}
    \label{fig:tc2}
\end{subfigure}
\hfill
\begin{subfigure}[t]{0.33\textwidth}
    \centering
    \includegraphics[width=\textwidth, keepaspectratio]{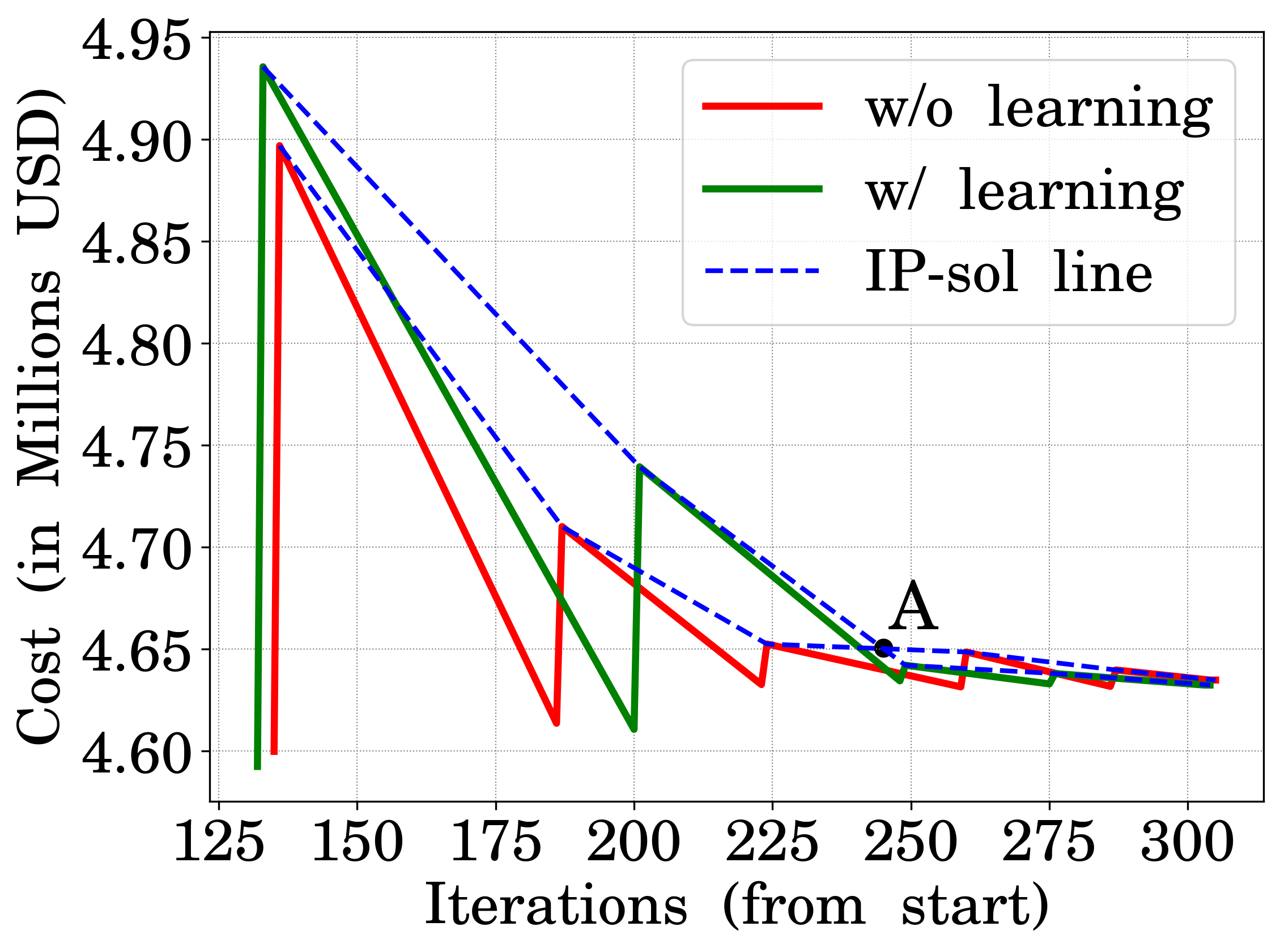}
    \caption{TC3}
    \label{fig:tc3}
\end{subfigure}
\caption{Plots of cost vs number of iterations for optimizer-runs with \& without the \textit{learning framework}}
\label{fig:runs}
\end{figure*}

\subsection{Hyperparameter Tuning}
Being in an assistive role, the learning model is tuned in a way to keep the run-time of learning-iterations as low as possible while ensuring a good $roc$ score ($\geq 0.9$). This in turn helps in keeping the optimizer's runtime under control. Two possible hyperparameters (epochs $n$ \& learning rate $\alpha$) are selected for hyperparameter tuning which is performed in two steps. In the first step, $n$ is varied as 100, 500 and 1000 with $\alpha$ varying as powers of 10 and its results are shown in Table~\ref{tab:tuning1}. Clearly, a reasonable choice is $n = 100$ with $\alpha = 0.01$ (highlighted in Table~\ref{tab:tuning1}). By selecting $n=100$ from here, a favorable baseline run-time is established for the second step in which $\alpha$ is then varied as multiples of the chosen power of 10. From its results, shown in Table~\ref{tab:tuning2}, the optimum value of $\alpha$ comes out as $0.03$. 

\subsection{\textit{AirCROP}-Runs with \textit{Learning Framework}}
For all three test-cases, the optimizer-runs are performed with \& without the \textit{learning framework} and the respective results are compared to establish the significance of the proposed contributions. These results are presented in Table~\ref{tab:runs}. In large-scale CPOPs, it is observed that the cost of integer solution at the end of the main optimization loop is extremely poor unless new pairings are also generated in the IP-phase. To overcome this limitation, the integer solution  from the main optimization loop is re-optimized using the same methodology (CG followed by IP). This re-optimization loop is repeated  multiple times till IP-cost becomes equal to its root LP-cost, or till a user defined cut-off. In the referred table, the values of cost and number of CG-iterations\footnote{modules other than CG are assumed in equivalence to 1 CG-iteration} $z$ for all optimization loops are presented for all test cases. It is observed that in optimizer-runs with the \textit{learning framework}, substantially lower costs are achieved, especially for TC1 \& TC3. These reductions are seen across datasets as well as successive optimization loops albeit there are some exceptions. Another critical observation is the achievement of lower-cost solution in fewer re-optimization loops while using the optimizer with the \textit{learning framework} for TC3.
\par Figures~\ref{fig:tc1}-\ref{fig:tc3} represent the comparison plots of cost vs number of iterations (from beginning of the runs) for all optimizer-runs and all datasets. The initial solutions are excluded from these figures as their respective costs are extremely poor in comparison to their respective optimal-costs, leading to distortion of scales. In Figures~\ref{fig:tc1} \&~\ref{fig:tc3}, the IP-sol. lines for runs with and without the \textit{learning framework} intersect at a point A. This implies that after point A, the run with the \textit{learning framework} provides better IP-solutions and that too in a shorter runtime. Moreover, each IP-solution is a sufficient solution in itself and the decision lies with the user to stop after any IP as per individual time constraints. The proposed \textit{learning framework} is only used to assist the CG-phase. Hence, the changes in IP-cost (mostly positive improvements) are attributed to the changes in the LP-solutions of their respective CG-phases. Even though a drop in costs is observed across the board, an increase (approximately doubled) in optimizer's runtime is observed when the \textit{learning framework} is used. This is attributed to the fact that training an auto-encoder is a time- and resource-intensive process.

\section{Conclusion} \label{sec:conclusion}

To the authors' knowledge, this research has proposed a first of its kind \textit{learning framework} within the paradigm of Airline Crew Pairing Optimization. This framework based on the Variational Graph Auto-Encoder, is shown to be capable of learning implicit combinatorial flight-connection patterns, on-the-fly, eventually leading up to significant cost savings within fewer CG-iterations. The efficacy and robustness of the proposed framework is endorsed through empirical results on three real-world, large and complex flight datasets of US-based airlines. On the flip side, the above advantages come at the expense of larger (almost double) run time for the overall optimizer. However, this runtime could be drastically reduced with the usage of GPU(s). Though the cost of computational resources may be significant, the corresponding cost-savings for an airline may well offset those in the long-run.

\section*{Acknowledgment}
This research work is an outcome of an Indo-Dutch joint research project. It is supported by the Ministry of Electronics and Information Technology (MEITY) from India [grant 13(4)/2015-CC\&BT], Netherlands Organization for Scientific Research (NWO) from the Netherlands, and General Electric (GE) Aviation. The authors would like to acknowledge
the invaluable support of GE Aviation team members: Saaju Paulose (Senior Manager), Arioli Arumugam (Senior
Director- Data \& Analytics), and Alla Rajesh (Senior Staff Data \& Analytics Scientist) for providing problem definition, real-world test cases, and for sharing domain-knowledge during numerous insightful discussions that helped
the authors in successfully completing this work.

\bibliographystyle{named}
\bibliography{ijcai20_v1}

\begin{thebibliography}{}

\bibitem[\protect\citeauthoryear{Aggarwal \bgroup \em et al.\egroup
  }{2018}]{aggarwal2018large}
Divyam Aggarwal, Dhish~Kumar Saxena, Michael Emmerich, and Saaju Paulose.
\newblock On large-scale airline crew pairing generation.
\newblock In {\em 2018 IEEE Symposium Series on Computational Intelligence
  (SSCI)}, pages 593--600. IEEE, 2018.

\bibitem[\protect\citeauthoryear{Aggarwal \bgroup \em et al.\egroup
  }{2020a}]{aggarwal2020aircrop}
Divyam Aggarwal, Dhish~Kumar Saxena, Thomas B{\"a}ck, and Michael Emmerich.
\newblock Aircrop: Airline crew pairing optimizer for complex flight networks
  involving multiple crew bases \& billion-plus variables.
\newblock {\em arXiv:2003.03994 [cs.MS]}, 2020.

\bibitem[\protect\citeauthoryear{Aggarwal \bgroup \em et al.\egroup
  }{2020b}]{aggarwal2020initializing}
Divyam Aggarwal, Dhish~Kumar Saxena, Thomas B{\"a}ck, and Michael Emmerich.
\newblock On initializing airline crew pairing optimization for large-scale
  complex flight networks.
\newblock {\em arXiv:2003.03994 [cs.AI]}, 2020.

\bibitem[\protect\citeauthoryear{Anbil \bgroup \em et al.\egroup
  }{1992}]{anbil1992global}
Ranga Anbil, Rajan Tanga, and Ellis~L. Johnson.
\newblock A global approach to crew-pairing optimization.
\newblock {\em IBM Systems Journal}, 31(1):71--78, 1992.

\bibitem[\protect\citeauthoryear{Anbil \bgroup \em et al.\egroup
  }{1998}]{anbil1998column}
Ranga Anbil, John~J Forrest, and William~R Pulleyblank.
\newblock Column generation and the airline crew pairing problem.
\newblock {\em Documenta Mathematica}, 3(1):677, 1998.

\bibitem[\protect\citeauthoryear{Aytug \bgroup \em et al.\egroup
  }{1994}]{aytug1994areview}
Haldun Aytug, Siddhartha Bhattacharyya, Gary Koehler, and Jane Snowdon.
\newblock A review of machine learning in scheduling.
\newblock {\em Engineering Management, IEEE Transactions on}, 41:165 -- 171, 06
  1994.

\bibitem[\protect\citeauthoryear{Barnhart \bgroup \em et al.\egroup
  }{1998}]{barnhart1998branch}
Cynthia Barnhart, Ellis~L Johnson, George~L Nemhauser, Martin~WP Savelsbergh,
  and Pamela~H Vance.
\newblock Branch-and-price: Column generation for solving huge integer
  programs.
\newblock {\em Operations research}, 46(3):316--329, 1998.

\bibitem[\protect\citeauthoryear{Barnhart \bgroup \em et al.\egroup
  }{2003}]{barnhart2003airline}
Cynthia Barnhart, Amy~M Cohn, Ellis~L Johnson, Diego Klabjan, George~L
  Nemhauser, and Pamela~H Vance.
\newblock Airline crew scheduling.
\newblock In {\em Handbook of transportation science}, pages 517--560.
  Springer, 2003.

\bibitem[\protect\citeauthoryear{Bengio \bgroup \em et al.\egroup
  }{2018}]{bengio2018machine}
Yoshua Bengio, Andrea Lodi, and Antoine Prouvost.
\newblock Machine learning for combinatorial optimization: a methodological
  tour d'horizon.
\newblock {\em CoRR}, abs/1811.06128, 2018.

\bibitem[\protect\citeauthoryear{Bernhard and
  Vygen}{2012}]{bernhard2008combinatorial}
Korte Bernhard and Jens Vygen.
\newblock {\em Combinatorial Optimization: Theory and Algorithms (5 ed.)}.
\newblock Springer, 2012.

\bibitem[\protect\citeauthoryear{Deb and
  Srinivasan}{2006}]{deb2006innovization}
Kalyanmoy Deb and Aravind Srinivasan.
\newblock Innovization: Innovating design principles through optimization.
\newblock In {\em Proceedings of the 8th annual conference on Genetic and
  evolutionary computation}, pages 1629--1636. ACM, 2006.

\bibitem[\protect\citeauthoryear{Desaulniers and
  Soumis}{2010}]{desaulniers2010airline}
G~Desaulniers and F~Soumis.
\newblock Airline crew scheduling by column generation.
\newblock {\em CIRRELT Spring School, Montr{\'e}al Canada}, 2010.

\bibitem[\protect\citeauthoryear{Gabriel~Crainic and
  Rousseau}{1987}]{gabriel1987column}
Teodor Gabriel~Crainic and Jean-Marc Rousseau.
\newblock The column generation principle and the airline crew scheduling
  problem.
\newblock {\em INFOR: Information Systems and Operational Research},
  25(2):136--151, 1987.

\bibitem[\protect\citeauthoryear{Gaur and Deb}{2016}]{gaur2016adaptive}
Abhinav Gaur and Kalyanmoy Deb.
\newblock Adaptive use of innovization principles for a faster convergence of
  evolutionary multi-objective optimization algorithms.
\newblock In {\em Proceedings of the 2016 on Genetic and Evolutionary
  Computation Conference Companion}, pages 75--76. ACM, 2016.

\bibitem[\protect\citeauthoryear{Hastie \bgroup \em et al.\egroup
  }{2009}]{hastie2009elements}
Trevor Hastie, Robert Tibshirani, and Jerome Friedman.
\newblock {\em The elements of statistical learning: data mining, inference,
  and prediction}.
\newblock Springer Science \& Business Media, 2009.

\bibitem[\protect\citeauthoryear{Khalil \bgroup \em et al.\egroup
  }{2017}]{khalil2017learning}
Elias Khalil, Hanjun Dai, Yuyu Zhang, Bistra Dilkina, and Le~Song.
\newblock Learning combinatorial optimization algorithms over graphs.
\newblock In {\em Advances in Neural Information Processing Systems}, pages
  6348--6358, 2017.

\bibitem[\protect\citeauthoryear{Kipf and
  Welling}{2016a}]{DBLP:journals/corr/KipfW16}
Thomas~N. Kipf and Max Welling.
\newblock Semi-supervised classification with graph convolutional networks.
\newblock {\em CoRR}, abs/1609.02907, 2016.

\bibitem[\protect\citeauthoryear{Kipf and
  Welling}{2016b}]{Kipf2016VariationalGA}
Thomas~N. Kipf and Max Welling.
\newblock Variational graph auto-encoders.
\newblock {\em ArXiv}, abs/1611.07308, 2016.

\bibitem[\protect\citeauthoryear{Land and Doig}{1960}]{land1960automatic}
AH~Land and AG~Doig.
\newblock An automatic method of solving discrete programming problems.
\newblock {\em Econometrica}, 28(3):497--520, 1960.

\bibitem[\protect\citeauthoryear{Li \bgroup \em et al.\egroup
  }{2018}]{li2018combinatorial}
Zhuwen Li, Qifeng Chen, and Vladlen Koltun.
\newblock Combinatorial optimization with graph convolutional networks and
  guided tree search.
\newblock In {\em Advances in Neural Information Processing Systems}, pages
  539--548, 2018.

\bibitem[\protect\citeauthoryear{L{\"u}bbecke}{2010}]{lubbecke2010column}
Marco~E L{\"u}bbecke.
\newblock Column generation.
\newblock {\em Wiley Encyclopedia of Operations Research and Management
  Science, John Wiley and Sons, Chichester, UK}, 2010.

\bibitem[\protect\citeauthoryear{Murty}{1983}]{MR720547}
Katta~G. Murty.
\newblock {\em Linear programming}.
\newblock John Wiley \& Sons, Inc., New York, 1983.
\newblock With a foreword by George B. Dantzig.

\bibitem[\protect\citeauthoryear{Priore \bgroup \em et al.\egroup
  }{2014}]{priore2014dynamic}
Paolo Priore, Alberto G{\'o}mez, Ra{\'u}l Pino, and Rafael Rosillo.
\newblock Dynamic scheduling of manufacturing systems using machine learning:
  An updated review.
\newblock {\em AI EDAM}, 28(1):83--97, 2014.

\bibitem[\protect\citeauthoryear{Soumis \bgroup \em et al.\egroup
  }{2019}]{Soumis2019FlightconnectionPF}
François Soumis, Yassine Yaakoubi, and Simon Lacoste-Julien.
\newblock Flight-connection prediction for airline crew scheduling to construct
  initial clusters for or optimizer.
\newblock 2019.

\bibitem[\protect\citeauthoryear{Vance \bgroup \em et al.\egroup
  }{1997}]{vance1997heuristic}
Pamela~H Vance, Cynthia Barnhart, Eric Gelman, Ellis~L Johnson, Alamuru
  Krishna, Deepa Mahidhara, George~L Nemhauser, and Ranjit Rebello.
\newblock A heuristic branch-and-price approach for the airline crew pairing
  problem.
\newblock Technical report lec-97-06, Georgia Institute of Technology, Atlanta,
  1997.

\bibitem[\protect\citeauthoryear{Wang \bgroup \em et al.\egroup
  }{2016}]{wang2016applying}
Peng Wang, Gang Zhao, and Xingren Yao.
\newblock Applying back-propagation neural network to predict bus traffic.
\newblock In {\em 2016 12th International Conference on Natural Computation,
  Fuzzy Systems and Knowledge Discovery (ICNC-FSKD)}, pages 752--756. IEEE,
  2016.

\bibitem[\protect\citeauthoryear{Zeren and {\"O}zkol}{2016}]{zeren2016novel}
Bahad{\i}r Zeren and Ibrahim {\"O}zkol.
\newblock A novel column generation strategy for large scale airline crew
  pairing problems.
\newblock {\em Expert Systems with Applications}, 55:133--144, 2016.

\end{thebibliography}

\end{document}